\documentclass{article} 
\usepackage[final]{colm2026_conference}
\usepackage{microtype}
\usepackage{hyperref}
\usepackage{url}
\usepackage{booktabs}

\usepackage{graphicx}
\usepackage{booktabs}

\usepackage{multirow}
\usepackage{subcaption}
\usepackage{wrapfig}

\usepackage{lineno}

\definecolor{darkblue}{rgb}{0, 0, 0.5}
\hypersetup{colorlinks=true, citecolor=darkblue, linkcolor=darkblue, urlcolor=darkblue}

\title{Multilingual Embedding Probes Fail to Generalize Across Learner Corpora}

\author{Laurits Lyngbaek$^*$, Ross Deans Kristensen-McLachlan \\
Aarhus University, Denmark \\
\textbf{Correspondence}: \texttt{laurits@cc.au.dk}$^*$
}

\begin{document}

\ifcolmsubmission
\linenumbers
\fi

\maketitle

\begin{abstract}
Do multilingual embedding models encode a language-general representation of proficiency? We investigate this by training linear and non-linear probes on hidden-state activations from seven embedding models (0.3-8B) to predict CEFR proficiency levels from learner texts across nine corpora and seven languages. We compare five probing architectures against a baseline trained on surface-level text features. Under in-distribution evaluation, probes achieve strong performance (Quadratic Weighted Kappa $\approx0.7$), substantially outperforming the surface baseline, with middle layers consistently yielding the best predictions. However, in cross-corpus evaluation performance collapses across all probe types and model sizes. Residual analysis reveals that out-of-distribution probes converge towards predicting uniformly distributed labels, indicating that the learned mappings capture corpus-specific distributional properties (topic, language, task type, rating methodology) rather than an abstract, transferable proficiency dimension. These results suggest that current multilingual embeddings do not straightforwardly encode language-general proficiency, with implications for representation-based approaches to proficiency-adaptive language technology.

\textbf{GitHub}: \href{https://github.com/INTERACT-LLM/proficiency-probing}{\texttt{https://github.com/INTERACT-LLM/proficiency-probing}}
\end{abstract}

\section{Introduction}

A growing body of work explores large language models as tools for supporting second-language learning, with potential benefits ranging from interactive practice to personalized instruction \citep{Han_2024}. A prerequisite for these systems is the ability to reliably represent and adapt to learner proficiency level \citep{DEVORE2025101745, jin-etal-2026-toward}. An obvious solution to this problem would be to rely primarily on prompt engineering, instructing an LLM to constrain its output to a given linguistic level.
However, recent work has shown that such prompting can be brittle, with proficiency level differentiation eroding over time as a result of so-called \textbf{alignment drift} \citep{almasi-kristensen-mclachlan-2025-alignment, jin-etal-2026-toward}. 
In these contexts, text generation models show a systematic tendency to move towards mid-range proficiency levels, making them unsuited for either beginners or more advanced learners \citep{benedetto2025assessing}.

These limitations motivate a shift towards analyses at the representation level. If multilingual models encode proficiency-relevant structure in their internal representations, this structure could in principle be leveraged for downstream applications, such as activation steering of generative models \citep{subramani-etal-2022-extracting, nguyen-etal-2025-multi, stolfo2025improving}. However, it is unclear whether such structure exists in a form that generalizes beyond the specific corpora used to identify it. Linguistic proficiency necessarily involves a wide range of lexical, syntactic, discursive and pragmatic features commonly targeted by probing studies \citep{li2023inferencetime, marks2024the, hollinsworth-etal-2024-language}, suggesting that aspects of proficiency and complexity should be recoverable from models. Nevertheless, proficiency is also a socially mediated process and there are multiple (sometimes contradictory) definitions of what linguistic proficiency really is, with a general consensus that proficiency is a multidimensional phenomenon \citep{housen-kuiken-2009}. It is therefore an open question whether standard probing approaches can transfer to a continuous, graded, and fuzzy concept such as proficiency.

\begin{figure}[t]
\begin{center}
\includegraphics[width=1\linewidth]{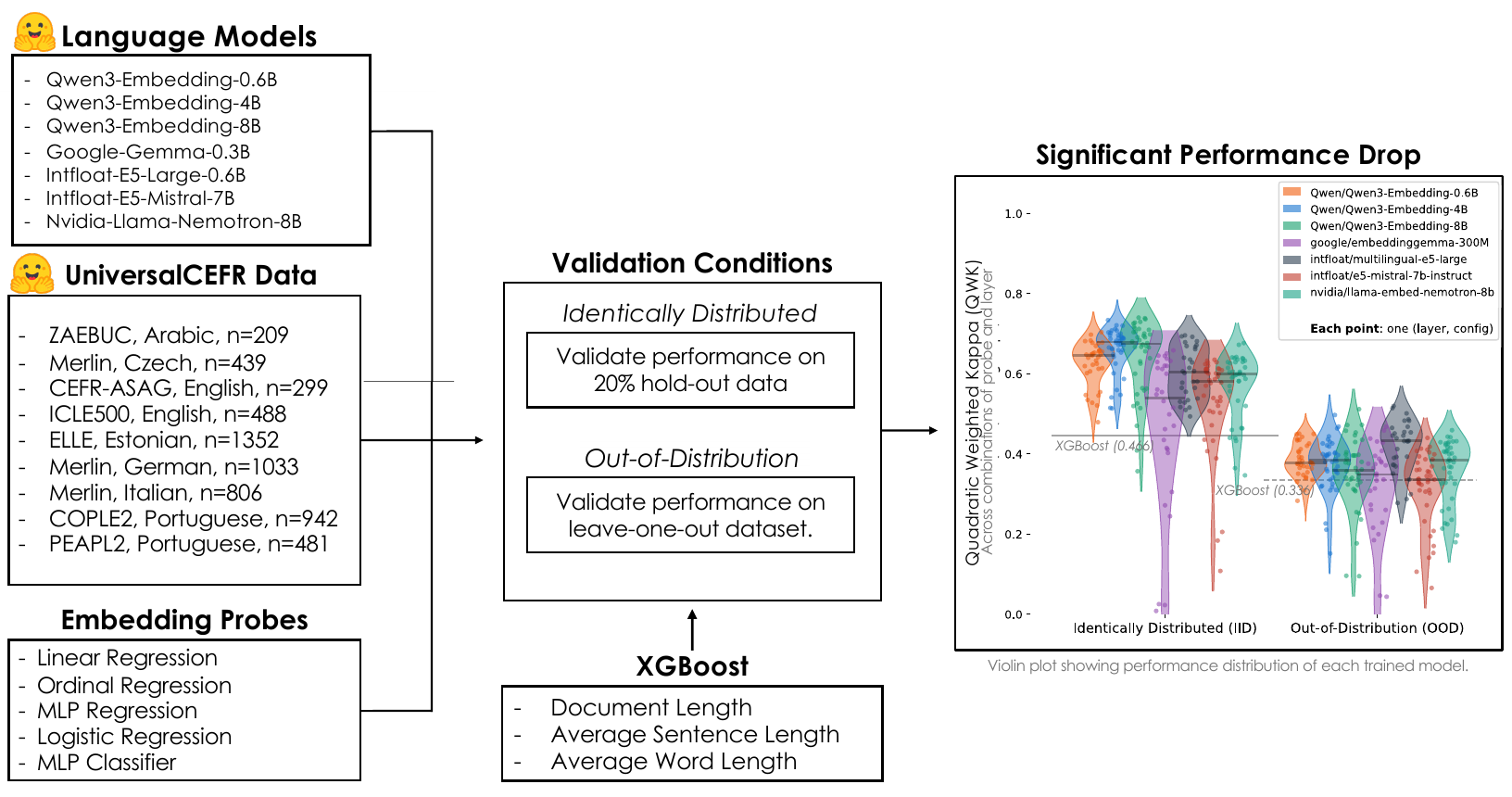}
\end{center}
\caption{Visualization of experimental design. All probes are used on every fifth hidden layers of each language model. In the Out-of-Distribution condition, this process is repeated for every dataset, by leaving it out from the training data, and afterwards testing performance on the held-out data.}
\end{figure}

In this paper, we investigate whether multilingual embedding models do in fact encode a language-general representation of proficiency which is recoverable via probing. 
We extract hidden-state activations across all layers for seven LLMs with varying architectures and training data.
We then train five different probe architectures on a corpus of learner texts labelled with expected proficiency level, with texts drawn from nine different corpora spanning seven languages \citep{imperial2025universalcefr}. Each probe architecture encodes different assumptions about the geometry of the proposed proficiency representation, allowing for both linear and non-linear structures.

We evaluate these probes under two conditions: an in-distribution split, where train and test data are drawn from the same data mixture; and a leave-one-out cross-corpus split, where an entire corpus is withheld from training. For in-distribution evaluation, probes substantially outperform a XGBoost baseline trained on surface features, achieving a Quadratic Weighted Kappa (QWK) $\approx0.73$ in middle layers of the Qwen3-4B model. However, during cross-corpus out-of-distribution evaluation, performance collapses uniformly across all probe architectures, model sizes, and layer depths. Residual analysis reveals that OOD predictions degenerate toward increasingly uniformly distributed labels, indicating that the probes have learned corpus-specific distributional regularities rather than a transferable proficiency dimension.

These findings contribute to the broader understanding of what high-level linguistic properties are and are not recoverable from pretrained representations \citep{hewitt-liang-2019-designing, hewitt-manning-2019-structural, chi-etal-2020-finding, belinkov-2022-probing, chang2022geometry, park2025the}. While probing has been shown to extract syntactic, semantic, and even some discourse-level properties from language models, our results suggest that proficiency resists the same treatment, at least in terms of cross-lingual transfer across corpora. Taken together with previous research into the alignment drift phenomenon, these results indicate that neither prompting nor representation extraction provides a straightforward path to adaptive, proficiency-aware language technology without careful attention to the distributional properties of training data and the confounds inherent in learner corpora.

\section{Related Work}
Recent related work on scoring of linguistic proficiency in written text is largely divided into two groups, the computational Automatic Essay Scoring (AES) and applied Second Language Acquisition (SLA) research.

\textbf{Automatic Essay Scoring (AES).} 
Automatic Essay Scoring is a branch of research investigating the computational challenge of converting learner produced essays into appropriate scores, for a systematic literature review see \citet{ramesh2022automated}.
As shown in the literature review, a high proportion of these works utilize the Automated Student Assessment Prize (ASAP) datasets \citep{shermis2014state, crossley2025large, mathias2018asap++}, trying to predict a \textit{holistic score} across 7 prompts. A highly related contemporary work from the AES literature is \citet{chi2026activations}, a paper predicting scores by using probes fitted to activations of attention heads. They achieve a Quadratic Weighted Kappa (QWK) $\approx0.67$ across a prompt-wise cross-validation, outperforming contemporary models \citep{li2025kaes, chen2024plaes, chen2023pmaes, do2023prompt, ridley2021automated, ridley2020prompt}. While these models do not display the same tendency of cross-validation loss, we note that all prompts are sampled from similar distributions of American students.

\textbf{Second Language Acquisition (SLA).} Proficiency prediction in SLA research is in contrast a field that investigate capabilities of models predicting ecologically sampled learner data. These models include algorithms using handcrafted linguistic features \citep{ forti2020malt,santucci2020automatic,  vajjala2018experiments, tack-etal-2017-human, vajjala2014automatic}, finetuning and probing of embedding models \citep{schmalz2021automatic, lagutina2024text, ahlers2025classifying} or in-context prediction via text-generation from decoder models \citep{benedetto2025assessing, mizumoto2023exploring, yancey2023rating, zhang2026automated, ahlers2025classifying}. Most of this research restricts its scope to a single dataset or language. This leaves a gap of work attempting to develop a generalizable multilingual proficiency model.
\section{Methods}
\subsection{Corpus and Input Structure}
The data for this study have been sampled from the UniversalCEFR dataset \citep{imperial2025universalcefr}, a large-scale, open, multilingual collection of 505,807 texts curated from educational and learner-oriented resources. UniversalCEFR aggregates 26 individual corpora spanning 13 languages including high-resource languages such as English, Spanish, French, and German, as well as mid- and low-resource languages such as Arabic, Estonian, Hindi, and Welsh. Texts are annotated with metadata related to the the proficiency level of the learner and covers all six levels available in the Common European Framework of Reference (CEFR). Texts are drawn from two production categories (learner text and reference text) and annotated at multiple granularities (sentence, paragraph, document, and dialogue level). 
The sub-corpora used for this experiment were selected based on being open source and consisting of learner-produced text. 
\begin{itemize}
\item \textbf{ZAEBUC} is a corpus of Arabic essays from first year Arabic-English bilingual university students, at Zayed University in the United Arab Emirates. The CEFR ratings are determined as the rounded average of three raters. \citep{habash-palfreyman-2022-zaebuc}
\item \textbf{MERLIN} corpus consists of three languages, Czech, German, and Italian. The data is compiled from standardized CEFR-related tests of L2 German and Italian at telc institute, Frankfurt and Czech at ÚJOP Institute, Prague. \citep{boyd-etal-2014-merlin}
\item \textbf{CEFR-ASAG} is a corpus of English essays by English-French bilingual university students at an unspecified university. The CEFR ratings are determined as the rounded average of three raters. \citep{tack-etal-2017-human, vajjala2018experiments}
\item \textbf{ICLE\_500} is a corpus of 500 English essays written by learners whose first-language was french, all answering the same prompt, originally stemming from the \textit{ETS Corpus of Non-Native Written English}. \citep{blanchard2014ets} The CEFR ratings were completed by crowd-sourcing 37 competent L1 judges, that do a pairwise ranking of the texts, the rankings of the essays are then converted to CEFR ratings. \citep{thwaites2025crowdsourced}
\item \textbf{ELLE} dataset is a corpus of Estonian learner essays, from the online learning platform Estonian Language Learning and Analysis Environment\footnote{https://elle.tlu.ee/}. The ELLE dataset does not explicitly mention how CEFR ratings are determined. \citep{allkivi2024elle}
\item \textbf{COPLE2}, is a corpus of test essays of Portuguese learners, either conducted as regular tests at language schools or as a accreditation exams. \citep{mendes-etal-2016-cople2}
\item \textbf{PEAPL2} is a corpus of Portuguese learners attending the Portuguese for Foreigners course at University of Coimbra. The CEFR labels are created in accordance to the CEFR-aligned course they were attending. \citet{martins2019corpus, del-rio-gayo-etal-2018-portuguese}
\end{itemize}
The final distribution of CEFR labels result in a imbalanced dataset, with a overweight of intermediate proficiencies, and an under-representation of beginner and advanced proficiencies. The classes C1 and C2 were combined into a C+ label, as the C2 label was deemed too skewed as ICLE500 and COPLE2 accounted for 93.7\% of essays. An overview of the datasets and their distribution of CEFR-levels can be seen in Table \ref{table-dataset-cefr}.

\begin{table}[t]
\begin{center}
\begin{tabular}{llccccccc}
\toprule
\multicolumn{1}{c}{\bf Dataset} & \multicolumn{1}{c}{\bf Language} & \multicolumn{1}{c}{\bf A1} & \multicolumn{1}{c}{\bf A2} & \multicolumn{1}{c}{\bf B1} & \multicolumn{1}{c}{\bf B2} & \multicolumn{1}{c}{\bf C+} & \multicolumn{1}{c}{\bf Size}\\
\midrule
ZAEBUC    & Arabic     &   0 &   7 & 111 &  80 &  11  (11/0) &  209 \\
MERLIN-cs & Czech      &   1 & 188 & 165 &  81 &   4   (4/0) &  439 \\
CEFR-ASAG & English    &  18 &  59 & 113 &  74 &  35  (30/5) &  299 \\
ICLE500   & English    &   0 &   1 & 114 & 218 & 155 (91/64) &  488 \\
ELLE      & Estonian   &   0 & 272 & 478 & 344 & 258 (258/0) & 1352 \\
MERLIN-de & German     &  57 & 306 & 331 & 293 &  46  (42/4) & 1033 \\
MERLIN-it & Italian    &  29 & 381 & 394 &   2 &   0   (0/0) &  806 \\
COPLE2    & Portuguese & 236 & 236 & 163 & 163 & 144 (72/72) &  942 \\
PEPPL2    & Portuguese &  78 &  89 & 204 & 70 &   40  (40/0) &  481 \\
\bottomrule
\\
\textbf{Total:} & & 419 & 1539 & 2073 & 1325 & 693 (548/145) & 6049 \\
\bottomrule
\bottomrule
\end{tabular}
\end{center}
\caption{CEFR distributions of the used datasets. All datasets are open source and available on Hugging Face. All datasets contain essays written by learners of the specified language.}\label{table-dataset-cefr}
\end{table}

\subsection{Model Activations and Latent Representations}
We created embedding representations for the sampled essays by using seven language models ranging from 0.3 to 8 billion parameters (see \autoref{table-models}). The three Qwen3 models \citep{qwen3embedding} have consistent architecture and training data between models, leaving only the model size as a varying condition. The remaining models are instruction-tuned embedding models, that vary along other dimensions, such as attention mechanism, pooling strategy, and training objective. We use a minimally informative instruction structured as: \texttt{"Instruct: Assess the CEFR level of the following text.\textbackslash nQuery: \{Essay\}"}

Models with causal attention use a final [EOS] token to aggregate document-level information, whose hidden state is optimized via contrastive loss to encode semantic similarity conditioned on the instruction query \citep{qwen3embedding, wang2024improving}. Models with bidirectional attention instead apply mean pooling, averaging the activations of all tokens to form the document-level representation \citep{vera2025embeddinggemma, wang2024multilingual, wang2022text}. While only the last layer of the model is used in benchmarking and downstream tasks, we probe multiple layers, as research suggest that representations of concepts resides more clearly in earlier layers \citep{conneau2020emerging, chi2020finding, skean2024does}.

\begin{table}[h]
\begin{center}
\begin{tabular}{lrcccc}
\toprule
\multicolumn{1}{c}{\bf Model} & \multicolumn{1}{c}{\bf Size} & \multicolumn{1}{c}{\bf Attention} & \multicolumn{1}{c}{\bf Pooling} & \multicolumn{1}{c}{\bf Vector Size} & \multicolumn{1}{c}{\bf MMTEB} \\
\midrule
\href{https://huggingface.co/Qwen/Qwen3-0.6B}{Qwen3-Embedding} & 0.6B & Causal & EOS & 1024 & 64.34 \\
\href{https://huggingface.co/Qwen/Qwen3-Embedding-4B}{Qwen3-Embedding} & 4.0B & Causal & EOS & 2560 & 69.45 \\
\href{https://huggingface.co/Qwen/Qwen3-Embedding-8B}{Qwen3-Embedding} & 8.0B & Causal & EOS & 4096 & 70.58 \\
\href{https://huggingface.co/google/embeddinggemma-300m}{Google-Gemma} & 0.3B & Bi-directional & Mean & 768 & 54.31 \\
\href{https://huggingface.co/intfloat/multilingual-e5-large}{Intfloat-E5-Large} & 0.6B & Bi-directional & Mean & 1024 & 55.08 \\
\href{https://huggingface.co/intfloat/e5-mistral-7b-instruct}{Intfloat-E5-Mistral} & 7.0B & Causal & EOS & 4096 & 53.08 \\
\href{https://huggingface.co/nvidia/llama-embed-nemotron-8b}{llama-embed-nemotron} & 8.0B & Bi-directional & Mean & 4096 & 69.46 \\
\bottomrule
\end{tabular}
\end{center}
\caption{Overview of embedding models used, including architecture details and their mean task accuracy on the Massive Multilingual Text Embedding Benchmark (MMTEB) \citep{enevoldsen2025mmteb}.}
\label{table-models}
\end{table}

\subsection{Probes and Baseline}
The literature on representations in language model have strong accounts for both linear representations and manifold representations. The Linear Representation Hypothesis proposes that certain representations are encoded by a one-dimensional subspace in the LLM activations \citep{park2023linear, elhage2022toy}. A growing body of research advocates for a Manifold Hypothesis that only views representations as locally linear \citep{modell2025origins, gurnee2026models, engels_not_2025}. We attempt to accommodate for this uncertainty by fitting multiple types of probes. The different probes are all continuous, but vary in their assumptions of the latent representation space. We hence fit five probes to the hidden activations of the three models. The probes are smaller machine learning models that optimize to predict the CEFR level of an essay, based on the corresponding hidden activations of a larger frozen language model.  The implementation details regarding the probes and baseline can be found in \autoref{appendix:parameters}.

\begin{itemize}
\item As a baseline for predictive power we fit an \textbf{XGBoost} model on three surface level features: Total Essay Length, Average Sentence Length and Average Word Length. These features correlate with proficiency without causally explaining linguistic capabilities.
\item A \textbf{linear regression} probe offer a linearly interpretable prediction. Geometrically the linear regression finds the one-dimensional subspace that represents CEFR-level. 
\item An \textbf{ordinal regression} probe models a linear vector inside of the latent LLM embedding alike the linear regression. However, the ordinal regression does not assume that the distance between CEFR levels are equally spaced. In principle, this means it is able to model the fact that improvement needed to progress from $B1 \to B2$ might be larger than between $A1 \to A2$. During testing, the latent proficiency representation is transformed to classes by a learned threshold based conversion \citep{rennie2005loss}. 
\item The \textbf{logistic regression} probe extends the linear probe family by treating CEFR prediction as a multi-class classification problem rather than a regression task. This probe fits a linear representation of of each class likelihood, and produces a probability distribution over CEFR classes via a softmax output layer. This breaks the assumption that a high-level representation such as linguistic proficiency necessarily spans a single continuous manifold and allows for different proficiency levels to be located in non-connected clusters.

\item To accommodate for the Manifold Hypothesis, we also fit an \textbf{MLP regressor} that approximates a potentially non-linear representation of proficiency. The MLP regression consists of a feedforward neural network with a single continuous output value. It takes LLM-embeddings of size $n$ as the input layer and fits a single hidden layer of size $n$ to predict numeric conversion of CEFR-labels. The hidden layer serves as a mapping function that transforms the non-linear high dimensional representation into a one-dimensional linear representation.
\item Finally, we implement an \textbf{MLP classifier} with the same architecture as the MLP regressor except it treats proficiency prediction as a discrete classification problem. Instead of a continuous latent proficiency space, it predicts a soft-maxed probability distribution of classes, allowing the model to delimit a set of non-linear clusters that represents the activation of each unique CEFR level. 
\end{itemize}

\section{Experiments}
\subsection{Identical Distribution Split}
We conduct two probing experiment to investigate if large language models encode a representation of linguistic proficiency in their internal hidden activations. In the first experiment, we attempt to disentangle the proficiency representation from the written language and the essay theme by including a wide range of learner essays, where we assume the only common denominator between CEFR-levels is their indicators of linguistic proficiency. We test a suite of probes as predictors of CEFR-levels to find a working proxy for linguistic proficiency. This entails testing every combination of probe architecture and language model size for every fifth layer. This grid-search of optimal parameters results in fitting a total of 345 probes. During training we hold out 20\% of the data stratified by CEFR-level for testing in the identical distribution (IID) condition. 

\begin{figure}[t]
\begin{center}
\includegraphics[width=1\linewidth]{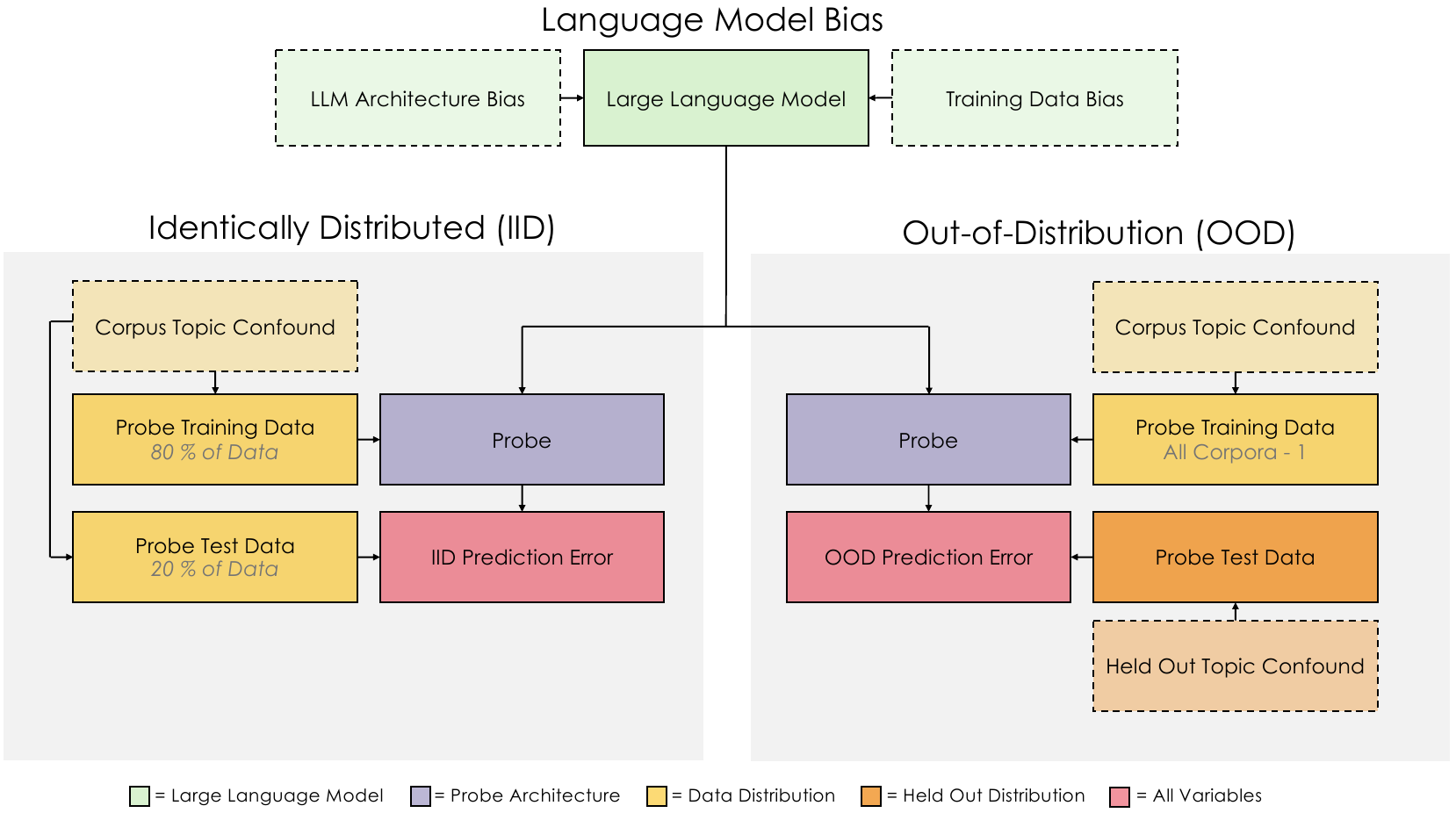}
\end{center}
\caption{Visualization of the potential sources of bias leading to an over-fitted proficiency representation. The OOD cross-validation ensures that the bias resulting from the corpus topic confound is represented in the probes predictive error.}
\label{fig:Bias}
\end{figure}

If our probe performs well on the hold-out data across languages and datasets, then we would normally conclude that our model has recovered a generalizable representation of proficiency.
This however rides on the assumption that our data is free of spurious correlations. This assumption is highly unlikely, as we can see from \autoref{table-dataset-cefr}, since every dataset has a unique label distribution, written language and task-prompts answered by learners. The potential overfitting may stem from multiple sources of biases as seen in \autoref{fig:Bias}. We therefore conduct an a secondary experiment.

\subsection{Out-of-Distribution, Leave-one-out Split}
In the second experiment, we test the probes' performance on Out-of-Distribution (OOD) data to disentangle the LLMs inherent training bias from the probes training bias. This is done by using a leave-one-out cross-corpus split, where an entire corpus is withheld from training. The theoretical reasoning for this is that the CEFR-datasets are often sampled from learning and examinations contexts, where the learner is given a exam prompt which instructs them asks them to write about a specific topic in a given style. As a result the IID condition have a high correlation between topic and CEFR-level, potentially allowing the machine learning algorithm to utilize the topic as a proxy for linguistic capabilities. The OOD condition does not resolve this problem during training but instead punishes predictive accuracy during testing. 

We train a model for every dataset at each grid search point in the IID condition, increasing the model count by factor of nine, resulting in a total of 2295 fitted probes. We test each probe on the corresponding held out dataset, results aggregated across layers and datasets are reported in \autoref{tab:results} alongside the IID results.

\section{Results and Discussion}
In this section we discuss the results of probing language models to find a linguistic proficiency representation. We argue that the results indicate a tendency for probes to overfit to spurious correlations of proficiency scores, such as language and topics.

\begin{table}[t]
\setlength{\tabcolsep}{4pt}
\begin{center}
\begin{tabular}{llcccc}
\toprule
\multicolumn{1}{c}{\bf Model} & \multicolumn{1}{c}{\bf Probe} & \multicolumn{1}{c}{\bf IID QWK} & \multicolumn{1}{c}{\bf IID F1} & \multicolumn{1}{c}{\bf OOD QWK} & \multicolumn{1}{c}{\bf OOD F1} \\
\midrule
XGB-surface &  & 0.446 (-) & 0.347 (-) & 0.336 (-) & 0.257 (-) \\
\midrule
\multirow{6}{*}{Qwen3-0.6B} & Linear Reg. & 0.682 (0.624) & 0.474 (0.435) & \textbf{0.451} \textbf{(0.401)} & \textbf{0.292} \textbf{(0.260)} \\
 & Ordinal Reg. & 0.677 (0.629) & 0.487 (0.447) & 0.450 (0.398) & 0.291 (0.255) \\
 & MLP Reg. & 0.698 (0.603) & 0.497 (0.404) & 0.435 (0.380) & 0.281 (0.255) \\
 & Logistic Reg. & 0.660 (0.626) & 0.549 (0.509) & 0.407 (0.363) & 0.290 (0.254) \\
 & MLP Clf. & \textbf{0.704} \textbf{(0.650)} & \textbf{0.584} \textbf{(0.534)} & 0.391 (0.377) & 0.264 \textbf{(0.260)} \\
 & {Mean} & \textit{0.684 (0.626)} & \textit{0.518 (0.466)} & \textit{0.427 (0.384)} & \textit{0.284 (0.257)} \\
\midrule
\multirow{6}{*}{Qwen3-4B} & Linear Reg. & 0.672 (0.630) & 0.437 (0.428) & 0.404 (0.323) & 0.249 (0.219) \\
 & Ordinal Reg. & 0.696 (0.665) & 0.455 (0.456) & 0.426 (0.365) & 0.282 (0.235) \\
 & MLP Reg. & 0.720 \textbf{(0.675)} & 0.537 (0.481) & 0.464 (0.395) & \textbf{0.297} (0.257) \\
 & Logistic Reg. & 0.703 (0.659) & 0.549 (0.524) & \textbf{0.497} (0.381) & 0.296 (0.252) \\
 & MLP Clf. & \textbf{0.734} (0.674) & \textbf{0.593} \textbf{(0.552)} & 0.468 \textbf{(0.396)} & 0.268 \textbf{(0.263)} \\
 & {Mean} & \textit{0.705 (0.661)} & \textit{0.514 (0.488)} & \textit{0.452 (0.372)} & \textit{0.278 (0.245)} \\
\midrule
\multirow{6}{*}{Qwen3-8B} & Linear Reg. & 0.677 (0.531) & 0.467 (0.359) & 0.377 (0.248) & 0.253 (0.194) \\
 & Ordinal Reg. & 0.692 (0.647) & 0.468 (0.437) & 0.458 (0.340) & 0.283 (0.227) \\
 & MLP Reg. & 0.726 (0.658) & 0.503 (0.466) & \textbf{0.512} (0.396) & \textbf{0.323} (0.262) \\
 & Logistic Reg. & 0.728 (0.683) & 0.569 (0.554) & 0.447 (0.372) & 0.306 (0.257) \\
 & MLP Clf. & \textbf{0.740} \textbf{(0.694)} & \textbf{0.602} \textbf{(0.566)} & 0.473 \textbf{(0.406)} & 0.311 \textbf{(0.275)} \\
 & {Mean} & \textit{0.712 (0.643)} & \textit{0.522 (0.476)} & \textit{0.453 (0.352)} & \textit{0.295 (0.243)} \\
 \midrule
\multirow{6}{*}{Gemma-0.3B} & Linear Reg. & \textbf{0.659} \textbf{(0.610)} & 0.471 (0.421) & \textbf{0.468} \textbf{(0.385)} & \textbf{0.316} \textbf{(0.268)} \\
 & Ordinal Reg. & 0.632 (0.571) & \textbf{0.476} \textbf{(0.444)} & 0.439 (0.358) & 0.288 (0.259) \\
 & MLP Reg. & 0.415 (0.175) & 0.288 (0.162) & 0.274 (0.127) & 0.193 (0.135) \\
 & Logistic Reg. & 0.632 (0.571) & 0.476 \textbf{(0.444)} & 0.439 (0.358) & 0.288 (0.259) \\
 & MLP Clf. & 0.555 (0.442) & 0.392 (0.346) & 0.379 (0.312) & 0.273 (0.227) \\
 & {Mean} & \textit{0.580 (0.475)} & \textit{0.407 (0.352)} & \textit{0.404 (0.311)} & \textit{0.272 (0.228)} \\
 \midrule
\multirow{6}{*}{E5-Large-0.6B} & Linear Reg. & 0.640 (0.588) & 0.457 (0.432) & \textbf{0.520} \textbf{(0.446)} & \textbf{0.345} \textbf{(0.301)} \\
 & Ordinal Reg. & 0.636 (0.581) & 0.469 (0.432) & 0.515 (0.438) & 0.339 (0.297) \\
 & MLP Reg. & 0.694 \textbf{(0.654)} & 0.524 (0.467) & 0.487 (0.423) & 0.318 (0.289) \\
 & Logistic Reg. & 0.581 (0.536) & 0.483 (0.417) & 0.433 (0.382) & 0.292 (0.271) \\
 & MLP Clf. & \textbf{0.696} (0.648) & \textbf{0.557} \textbf{(0.536)} & 0.451 (0.412) & 0.287 (0.282) \\
 & {Mean} &  \textit{0.649 (0.601)} &\textit{ 0.498 (0.457)} &\textit{ 0.481 (0.420)} & \textit{0.316 (0.288)} \\
 \midrule
\multirow{6}{*}{Mistral-7B} & Linear Reg. & 0.624 (0.459) & 0.434 (0.342) & 0.418 (0.285) & 0.273 (0.232) \\
 & Ordinal Reg. & 0.605 (0.509) & 0.407 (0.371) & \textbf{0.440} (0.325) & 0.289 (0.247) \\
 & MLP Reg. & 0.629 (0.579) & 0.443 (0.425) & 0.404 (0.328) & 0.278 \textbf{(0.256)} \\
 & Logistic Reg. & 0.633 (0.534) & 0.466 (0.442) & 0.408 (0.316) & \textbf{0.295} (0.249) \\
 & MLP Clf. & \textbf{0.635} \textbf{(0.582)} & \textbf{0.527} \textbf{(0.463)} & 0.399 \textbf{(0.332)} & 0.264 \textbf{(0.256)} \\
 & {Mean} & \textit{0.625 (0.533)} & \textit{0.455 (0.408)} & \textit{0.414 (0.317)} &\textit{ 0.280 (0.248)} \\
 \midrule
\multirow{6}{*}{Llama-8B} & Linear Reg. & 0.602 (0.486) & 0.403 (0.344) & 0.367 (0.279) & 0.266 (0.236) \\
 & Ordinal Reg. & 0.616 (0.538) & 0.512 (0.467) & 0.432 (0.382) & 0.293 (0.274) \\
 & MLP Reg. & 0.638 (0.598) & 0.479 (0.434) & 0.443 (0.367) & \textbf{0.303} (0.276) \\
 & Logistic Reg. & 0.641 (0.583) & 0.512 (0.467) & 0.423 (0.382) & 0.293 (0.274) \\
 & MLP Clf. & \textbf{0.677} \textbf{(0.629)} & \textbf{0.578} \textbf{(0.500)} & \textbf{0.459} \textbf{(0.403)} & 0.297 \textbf{(0.284)} \\
 & {Mean} & \textit{0.635 (0.572)} & \textit{0.479 (0.428)} & \textit{0.427 (0.360)} & \textit{0.292 (0.268)} \\
\bottomrule
\end{tabular}
\end{center}
\caption{QWK and macro-F1 results across model sizes and probe configurations. Values indicate performance at the best-performing layer. Values in parentheses indicate mean performance across all layers. IID = in-distribution; OOD = out-of-distribution.}
\label{tab:results}
\end{table}

\subsection{Interaction between Predictive Accuracy, Language Model and Probe}
An overview of our experimental results can be seen in \autoref{tab:results}. We both report the best scoring probe and average score of probes for each combination of LLM and probe. The average QWK score of all probes in the IID condition is 0.587. In comparison the average QWK score of probes in the OOD condition is 0.379. This is a conditional 37.9\% drop in performance when testing out of distribution. 
These drops in performance result in the IID probes on average achieving a 0.141 higher QWK than the baseline, while the OOD probes on average were 0.043 higher than the baseline.

\begin{figure}[t]
\begin{center}
\begin{subfigure}[t]{0.48\linewidth}
    \includegraphics[width=\linewidth]{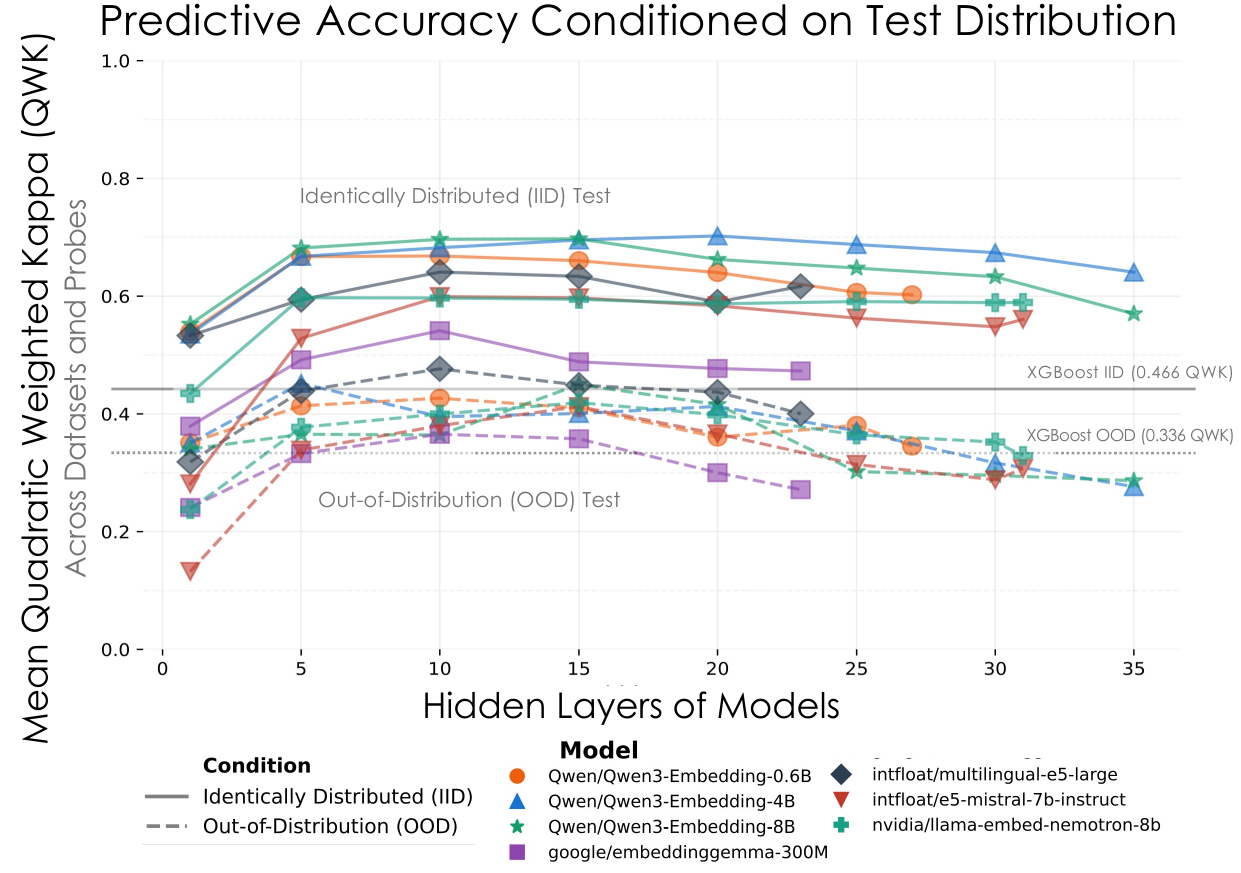}
    \caption{Predictive accuracy for the Qwen3 family of LLMs averaged across datasets and probes.}
    \label{fig:Models}
\end{subfigure}
\hfill
\begin{subfigure}[t]{0.48\linewidth}
    \includegraphics[width=\linewidth]{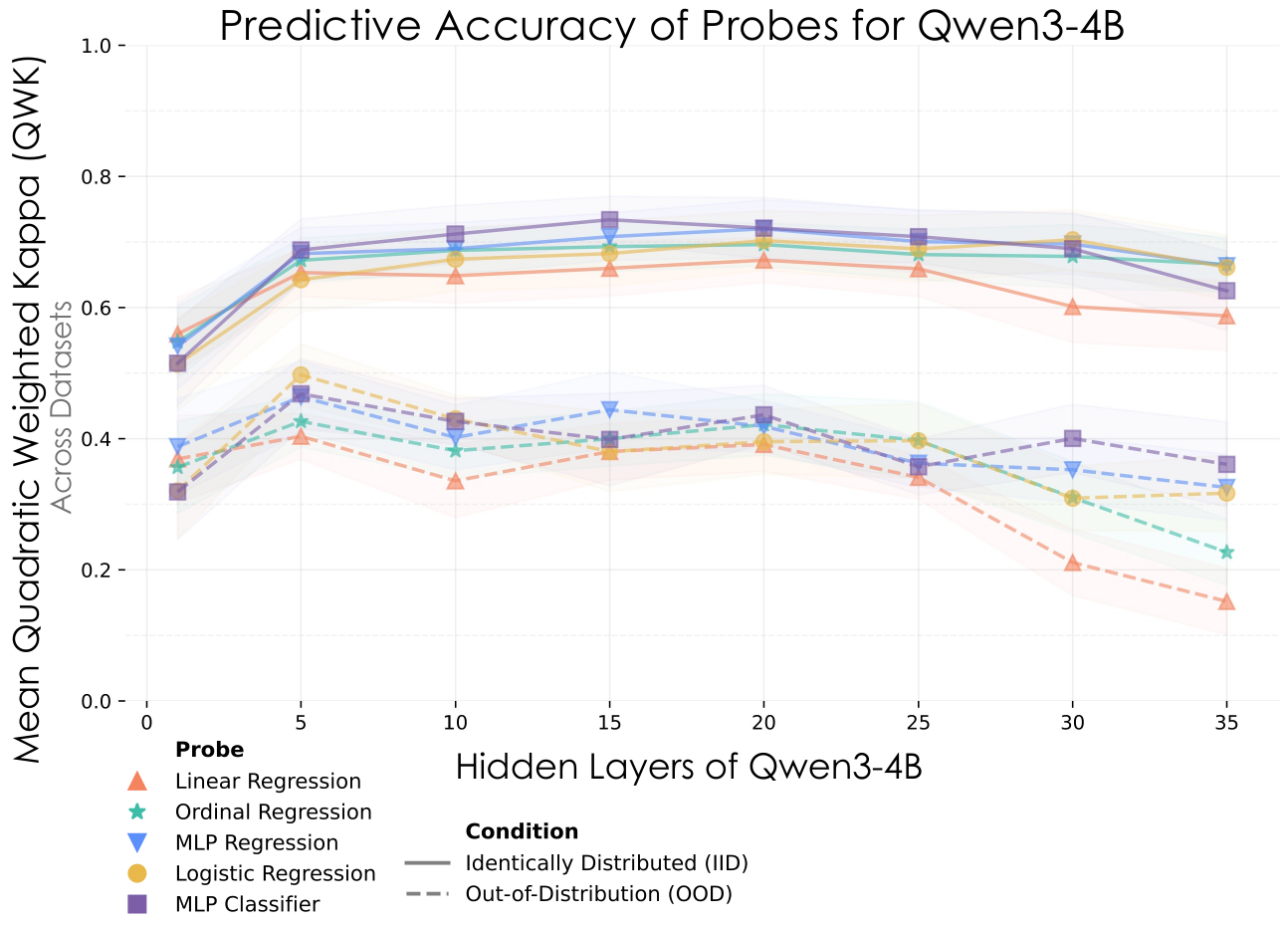}
    \caption{Predictive accuracy of probes for hidden layers of Qwen3-4B averaged across datasets. }
    \label{fig:Probes}
\end{subfigure}
\end{center}
\caption{Layer-wise trend of Quadratic Weighted Kappa of probes across hidden layers.}
\label{fig:both}
\end{figure}

As the language models grow in size, the general trend is toward more accurate proficiency representations. This is both true for the highlighted best probe and for the average of best probes. However, this pattern seems to break down when looking at average probe performance across all layers, where the Qwen-4B model appears to outperform the Qwen-8B model. This seems to be a result of the Linear Regression probe performing unusually poorly; excluding it, the 8B model marginally outperforms the 4B by 0.002. As shown in \autoref{fig:Models} this should be interpreted with the caveat that the difference between 4B and 8B is non-significant for all tested layers except layer 35.

\begin{figure}[htbp]
\begin{center}
\includegraphics[width=0.92\linewidth]{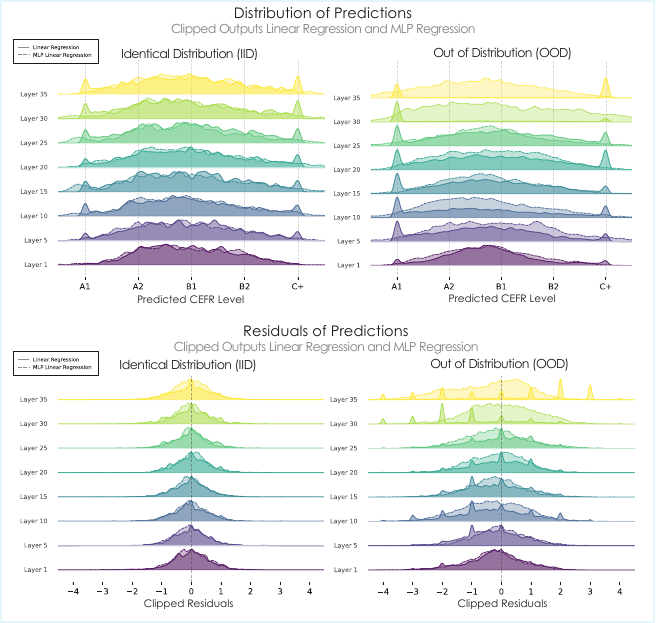}
\end{center}
\caption{Ridgeplot of predicted distribution (upper plot) and residuals of predictions (lower plot). Predictions are clipped to the closest real label if below A1 or above C+. The investigated probes are the linear regression and MLP regression on Qwen3-4B.}
\label{fig:Ridge}
\end{figure}

\autoref{fig:Models} is the first of three plots that investigate the effect of respectively model size, probe architecture, and datasets effect on the predictive capabilities of probes.  The figure displays the average QWK of probes fitted on the activations of every fifth hidden layer. While the IID condition confirms the hypothesized result that the intermediate layer of larger models represents the concept most accurately, the OOD test shows an unexpected loss of accuracy. Throughout the layers of the model, the model with the best performing average OOD probe fluctuates irregularly, making it impossible to conclude what model have the best representation of linguistic proficiency. While we could conclude from \autoref{tab:results} that a MLP regression probe on Qwen3-8B performs the best on unseen distributions, the lack of stability seen in \autoref{fig:Models} seem to indicate that this might not be the case for other distributions of learner essays.

To investigate the respective stability of the different probe architectures we create a similar visualization for the Qwen3-4B probes. The 4B model was chosen as the case study given it had the best mean aggregated probe performance. The results can be seen in \autoref{fig:Probes}, where the performance for each type of probe in both the IID and OOD condition is shown for their respective hidden layer. We once again see the same trend as in \autoref{fig:Models}, that the models are not able to generalize during the OOD condition. Another reoccurring trend seems to be that the performance in the OOD condition is a lot more unstable. The best performing probe architecture seems to fluctuate between hidden layers. A curious observation is that the linear probes seem to decrease in representation accuracy in the later layers, advocating for an interpretation that the model transforms its proficiency representation from linear to non-linear around layer 25.

The predictive failure of the linear probe is further investigated in \autoref{fig:Ridge}, where the layer-wise clipped predictions and residuals of the linear regression probe and MLP regression probe is compared for Qwen3-4B. We see that in the IID linear probe have tendency to occasionally predict outside of class-bounds, but still retain capabilities of detecting intermediate proficiency essays. 
This tendency is greatly exacerbated in the OOD condition, where it seems that after layer 25, no generalizable linear representations of proficiency is captured by the linear probe. A visualization of unclipped residuals for all layers of the Qwen3-4B linear regression probe is available in \autoref{appendix:raw_residuals}.

\begin{figure}[t]
\begin{center}
\includegraphics[width=1\linewidth]{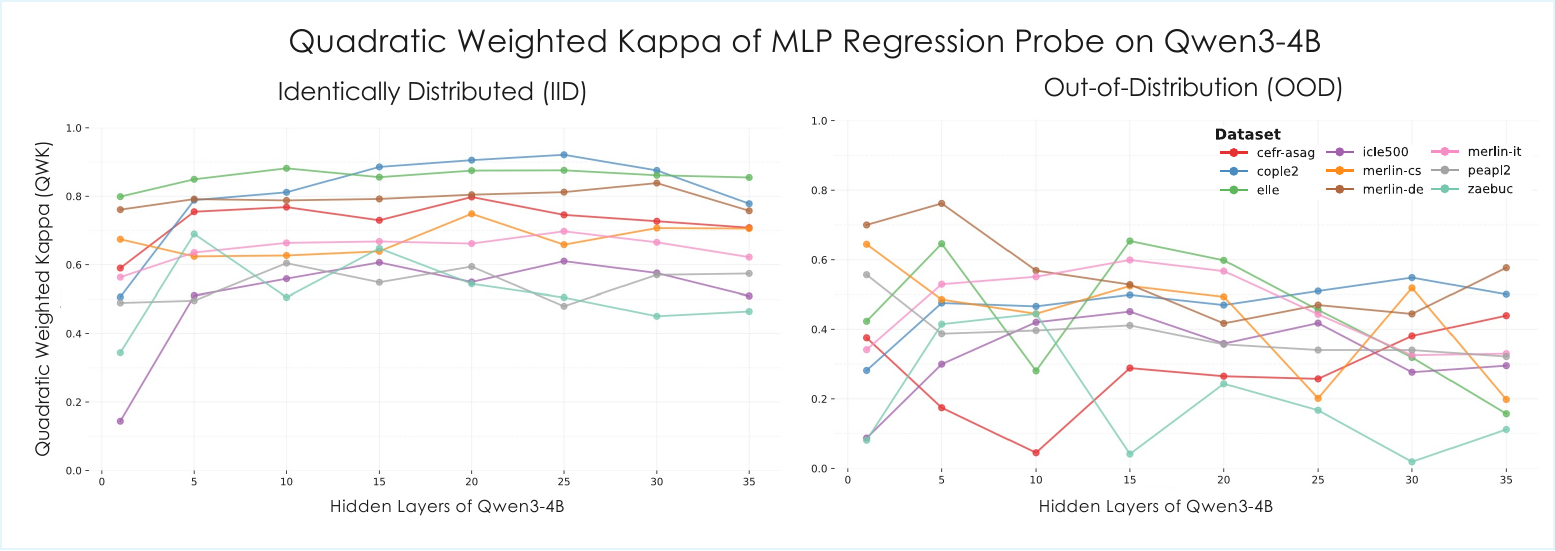}
\end{center}
\caption{Mean QWK for the Qwen3-4B MLP regression probe in the IID and OOD condition.}
\label{fig:Dataset}
\end{figure}

While the MLP regression is the best performing continuous probe for Qwen3-4B, it is still unclear if the OOD loss in performance is consistent across all datasets. We investigate this in \autoref{fig:Dataset}, where the performance of the MLP regression probe is visualized for each dataset across layers. The general trend of increased instability in the OOD condition reappears in this aggregation. 
Notably, the lowest QWK score in the final layer are ELLE and ZAEBUC, the only datasets comprising non Indo-European languages (Estonian and Arabic respectively). These languages exhibit the lowest OOD performance in the final layer of 4B model, indicating a potentially different proficiency representation in the model. While our language pool is not diverse enough to conclude whether this trend generalizes to more typologically distinct languages, it has worrying downstream implications, insofar as learning systems optimized with such representations would break in ways that disadvantage learners of these languages (see also our \hyperref[ssec:Ethics Statement]{Ethics Statement}).

\section{Conclusion}
Our results indicate that probing techniques of linguistic proficiency have a strong tendency to overfit corpus-specific distributional properties, such that they loose any meaningful interpretation when applied to non-training data.
This is inline with similar results which demonstrate that finetuning of encoder models for natural language inference generalize poorly to OOD-testing \citep{stacey2026improving}. These findings further challenge the predictive findings on related works that only test within distribution \citep{ahlers2025classifying, schmalz2021automatic}. These results warrant further investigation on how to mitigate bias from corpus-specific distributional properties during statistical representation discovery.

We hence conclude that that representation probing is not a reliable technique for discovering a language-general proficiency score. This is hypothesized to be a result of the probing algorithms overfitting to correlations between proficiency level and corpus-specific properties such as language and topic. Our results are hence in agreement with research which construes linguistic proficiency as a multidimensional phenomenon, which makes it seem unlikely that complexity as such can be fully disentangled from broader discursive properties such as register, genre, and style. If this is the case, this suggests that the \textit{steering vector} approach will not be enough, in and of itself, to create proficiency-adaptive computer-assisted language learning technologies.


\newpage
\section*{Ethics Statement}\label{ssec:Ethics Statement}

This study uses learner text data drawn exclusively from open, publicly available corpora. All datasets are redistributed under their original licences, and no new data collection involving human participants was conducted for this research.

The primary ethical concern raised by this work is the risk of deploying LLM-based computer assisted language learning systems without adequate understanding of their generalization limits. Our results demonstrate that linear probes trained on multilingual embeddings achieve strong in-distribution performance but fail systematically when applied to unseen learner corpora. A system built on this approach and deployed in an educational context could produce confident but misleading proficiency assessments for learners whose writing does not match the distributional properties of the training data. This risk is compounded by the opacity of embedding-based approaches, which may make such failures difficult for practitioners to detect. We therefore caution strongly against treating high in-distribution probe performance as evidence of deployable proficiency assessment capability.

While our dataset is multilingual, the majority of the languages included in our analysis are Indo-European, with only Arabic and Estonian coming from different language families (Semitic and Uralic, respectively). While our results demonstrate that complexity probes do not generalize within the IE language family, it remains unclear how much divergence would be seen if extended to even more languages.  Both of the non-IE languages exhibit the lowest OOD performance in the final layer of the Qwen3-4B model, indicating a potentially different proficiency representation in the model.  

The consistent under-performance of probes on these languages in the OOD condition suggests that current multilingual embedding models may encode proficiency-relevant information in ways that disadvantage learners of less-resourced or typologically distinct languages. Any downstream application of this approach should attend carefully to this disparity and we strongly encourage future research to consider typologically diverse language data.


\section*{Acknowledgments}
This work was partially supported by the Danish National Research Foundation (DNRF193) through TEXT: Center for Contemporary Cultures of Text, Aarhus University. 

\bibliography{colm2026_conference}

@article{qwen3embedding,
  title={Qwen3 Embedding: Advancing Text Embedding and Reranking Through Foundation Models},
  author={Zhang, Yanzhao and Li, Mingxin and Long, Dingkun and Zhang, Xin and Lin, Huan and Yang, Baosong and Xie, Pengjun and Yang, An and Liu, Dayiheng and Lin, Junyang and Huang, Fei and Zhou, Jingren},
  journal={arXiv preprint arXiv:2506.05176},
  year={2025}
}

@inproceedings{wang2024improving,
  title={Improving text embeddings with large language models},
  author={Wang, Liang and Yang, Nan and Huang, Xiaolong and Yang, Linjun and Majumder, Rangan and Wei, Furu},
  booktitle={Proceedings of the 62nd Annual Meeting of the Association for Computational Linguistics (Volume 1: Long Papers)},
  pages={11897--11916},
  year={2024}
}

@article{vera2025embeddinggemma,
  title={Embeddinggemma: Powerful and lightweight text representations},
  author={Vera, Henrique Schechter and Dua, Sahil and Zhang, Biao and Salz, Daniel and Mullins, Ryan and Panyam, Sindhu Raghuram and Smoot, Sara and Naim, Iftekhar and Zou, Joe and Chen, Feiyang and others},
  journal={arXiv preprint arXiv:2509.20354},
  year={2025}
}

@article{wang2024multilingual,
  title={Multilingual e5 text embeddings: A technical report},
  author={Wang, Liang and Yang, Nan and Huang, Xiaolong and Yang, Linjun and Majumder, Rangan and Wei, Furu},
  journal={arXiv preprint arXiv:2402.05672},
  year={2024}
}

@article{wang2022text,
  title={Text embeddings by weakly-supervised contrastive pre-training},
  author={Wang, Liang and Yang, Nan and Huang, Xiaolong and Jiao, Binxing and Yang, Linjun and Jiang, Daxin and Majumder, Rangan and Wei, Furu},
  journal={arXiv preprint arXiv:2212.03533},
  year={2022}
}

@article{babakhin2025llama,
  title={Llama-Embed-Nemotron-8B: A Universal Text Embedding Model for Multilingual and Cross-Lingual Tasks},
  author={Babakhin, Yauhen and Osmulski, Radek and Ak, Ronay and Moreira, Gabriel and Xu, Mengyao and Schifferer, Benedikt and Liu, Bo and Oldridge, Even},
  journal={arXiv preprint arXiv:2511.07025},
  year={2025}
}

@article{enevoldsen2025mmteb,
  title={Mmteb: Massive multilingual text embedding benchmark},
  author={Enevoldsen, Kenneth and Chung, Isaac and Kerboua, Imene and Kardos, M{\'a}rton and Mathur, Ashwin and Stap, David and Gala, Jay and Siblini, Wissam and Krzemi{\'n}ski, Dominik and Winata, Genta Indra and others},
  journal={arXiv preprint arXiv:2502.13595},
  year={2025}
}

@inproceedings{stacey2026improving,
  title={Improving the OOD Performance of Closed-Source LLMs on NLI Through Strategic Data Selection},
  author={Stacey, Joe and Alazraki, Lisa and Ubhi, Aran and Ermis, Beyza and Mueller, Aaron and Rei, Marek},
  booktitle={Findings of the Association for Computational Linguistics: EACL 2026},
  pages={5378--5404},
  year={2026}
}

@article{benedetto2025assessing,
  title={Assessing how accurately large language models encode and apply the common European framework of reference for languages},
  author={Benedetto, Luca and Gaudeau, Gabrielle and Caines, Andrew and Buttery, Paula},
  journal={Computers and Education: Artificial Intelligence},
  volume={8},
  pages={100353},
  year={2025},
  publisher={Elsevier}
}

@article{imperial2025universalcefr,
  title = {{UniversalCEFR: Enabling Open Multilingual Research on Language Proficiency Assessment}},
  author = {Joseph Marvin Imperial and Abdullah Barayan and Regina Stodden and Rodrigo Wilkens 
    and Ricardo Muñoz Sánchez and Lingyun Gao and Melissa Torgbi and Dawn Knight and Gail Forey 
    and Reka R. Jablonkai and Ekaterina Kochmar and Robert Reynolds and Eugénio Ribeiro and 
    Horacio Saggion and Elena Volodina and Sowmya Vajjala and Thomas François and 
    Fernando Alva-Manchego and Harish Tayyar Madabushi},
  journal = {arXiv preprint arXiv:2506.01419},
  year = {2025},
  url = {https://arxiv.org/abs/2506.01419}}

@inproceedings{habash-palfreyman-2022-zaebuc,
    title = "{ZAEBUC}: An Annotated {A}rabic-{E}nglish Bilingual Writer Corpus",
    author = "Habash, Nizar  and
      Palfreyman, David",
    editor = "Calzolari, Nicoletta  and
      B{\'e}chet, Fr{\'e}d{\'e}ric  and
      Blache, Philippe  and
      Choukri, Khalid  and
      Cieri, Christopher  and
      Declerck, Thierry  and
      Goggi, Sara  and
      Isahara, Hitoshi  and
      Maegaard, Bente  and
      Mariani, Joseph  and
      Mazo, H{\'e}l{\`e}ne  and
      Odijk, Jan  and
      Piperidis, Stelios",
    booktitle = "Proceedings of the Thirteenth Language Resources and Evaluation Conference",
    month = jun,
    year = "2022",
    address = "Marseille, France",
    publisher = "European Language Resources Association",
    url = "https://aclanthology.org/2022.lrec-1.9/",
    pages = "79--88"
}

@inproceedings{boyd-etal-2014-merlin,
    title = "The {MERLIN} corpus: Learner language and the {CEFR}",
    author = {Boyd, Adriane  and
      Hana, Jirka  and
      Nicolas, Lionel  and
      Meurers, Detmar  and
      Wisniewski, Katrin  and
      Abel, Andrea  and
      Sch{\"o}ne, Karin  and
      {\v{S}}tindlov{\'a}, Barbora  and
      Vettori, Chiara},
    editor = "Calzolari, Nicoletta  and
      Choukri, Khalid  and
      Declerck, Thierry  and
      Loftsson, Hrafn  and
      Maegaard, Bente  and
      Mariani, Joseph  and
      Moreno, Asuncion  and
      Odijk, Jan  and
      Piperidis, Stelios",
    booktitle = "Proceedings of the Ninth International Conference on Language Resources and Evaluation ({LREC}'14)",
    month = may,
    year = "2014",
    address = "Reykjavik, Iceland",
    publisher = "European Language Resources Association (ELRA)",
    url = "https://aclanthology.org/L14-1488/",
    pages = "1281--1288"
}

@inproceedings{tack-etal-2017-human,
    title = {Human and Automated {CEFR}-based Grading of Short Answers},
    author = {Tack, Ana{\"\i}s and Fran{\c{c}}ois, Thomas and Roekhaut, Sophie and Fairon, C{\'e}drick},
    booktitle = {Proceedings of the 12th Workshop on Innovative Use of {NLP} for Building Educational Applications},
    month = sep,
    year = {2017},
    address = {Copenhagen, Denmark},
    publisher = {Association for Computational Linguistics},
    url = {https://aclanthology.org/W17-5018},
    doi = {10.18653/v1/W17-5018},
    pages = {169--179}
}

@article{thwaites2025crowdsourced,
  title={Crowdsourced comparative judgement for evaluating learner texts: How reliable are judges recruited from an online crowdsourcing platform?},
  author={Thwaites, Peter and Vandeweerd, Nathan and Paquot, Magali},
  journal={Applied Linguistics},
  volume={46},
  number={4},
  pages={611--628},
  year={2025},
  publisher={Oxford University Press UK}
}

@article{blanchard2014ets,
  title={ETS Corpus of non-native written English LDC2014T06},
  author={Blanchard, Daniel and Tetreault, Joel and Higgins, Derrick and Cahill, Aoife and Chodorow, Martin},
  journal={Philadelphia: Linguistic Data Consortium},
  year={2014}
}

@article{allkivi2024elle,
  title={ELLE--estonian language learning and analysis environment},
  author={Allkivi, Kais and Eslon, Pille and Kamarik, Taavi and Kert, Karina and Kippar, Jaagup and Kodasma, Harli and Maine, Silvia and Norak, Kaisa},
  journal={Baltic Journal of Modern Computing},
  volume={12},
  number={4},
  pages={560--569},
  year={2024},
  publisher={University of Latvia}
}

@inproceedings{vajjala2018experiments,
  title={Experiments with universal CEFR classification},
  author={Vajjala, Sowmya and Rama, Taraka},
  booktitle={Proceedings of the thirteenth workshop on innovative use of NLP for building educational applications},
  pages={147--153},
  year={2018}
}

@inproceedings{mendes-etal-2016-cople2,
    title = "The {COPLE}2 corpus: a learner corpus for {P}ortuguese",
    author = "Mendes, Am{\'a}lia  and
      Antunes, Sandra  and
      Janssen, Maarten  and
      Gon{\c{c}}alves, Anabela",
    editor = "Calzolari, Nicoletta  and
      Choukri, Khalid  and
      Declerck, Thierry  and
      Goggi, Sara  and
      Grobelnik, Marko  and
      Maegaard, Bente  and
      Mariani, Joseph  and
      Mazo, Helene  and
      Moreno, Asuncion  and
      Odijk, Jan  and
      Piperidis, Stelios",
    booktitle = "Proceedings of the Tenth International Conference on Language Resources and Evaluation ({LREC}'16)",
    month = may,
    year = "2016",
    address = "Portoro{\v{z}}, Slovenia",
    publisher = "European Language Resources Association (ELRA)",
    url = "https://aclanthology.org/L16-1511/",
    pages = "3207--3214"
}

@article{martins2019corpus,
  title={Corpus de produ{\c{c}}{\~o}es escritas de aprendentes de PL2 (PEAPL2): Subcorpus Portugu{\^e}s l{\'\i}ngua estrangeira},
  author={Martins, Cristina and Ferreira, T and Sitoe, M and Abrantes, C and Janssen, M and Fernandes, A and Silva, A and Lopes, I and Pereira, I and Santos, J},
  journal={Coimbra: CELGA-ILTEC},
  year={2019}
}

@inproceedings{del-rio-gayo-etal-2018-portuguese,
    title = "A {P}ortuguese Native Language Identification Dataset",
    author = "del R{\'i}o Gayo, Iria  and
      Zampieri, Marcos  and
      Malmasi, Shervin",
    editor = "Tetreault, Joel  and
      Burstein, Jill  and
      Kochmar, Ekaterina  and
      Leacock, Claudia  and
      Yannakoudakis, Helen",
    booktitle = "Proceedings of the Thirteenth Workshop on Innovative Use of {NLP} for Building Educational Applications",
    month = jun,
    year = "2018",
    address = "New Orleans, Louisiana",
    publisher = "Association for Computational Linguistics",
    url = "https://aclanthology.org/W18-0534/",
    doi = "10.18653/v1/W18-0534",
    pages = "291--296"
}

@article{park2023linear,
  title={The linear representation hypothesis and the geometry of large language models},
  author={Park, Kiho and Choe, Yo Joong and Veitch, Victor},
  journal={arXiv preprint arXiv:2311.03658},
  year={2023}
}

@article{elhage2022toy,
  title={Toy models of superposition},
  author={Elhage, Nelson and Hume, Tristan and Olsson, Catherine and Schiefer, Nicholas and Henighan, Tom and Kravec, Shauna and Hatfield-Dodds, Zac and Lasenby, Robert and Drain, Dawn and Chen, Carol and others},
  journal={arXiv preprint arXiv:2209.10652},
  year={2022}
}

@article{modell2025origins,
  title={The origins of representation manifolds in large language models},
  author={Modell, Alexander and Rubin-Delanchy, Patrick and Whiteley, Nick},
  journal={arXiv preprint arXiv:2505.18235},
  year={2025}
}

@article{gurnee2026models,
  title={When models manipulate manifolds: The geometry of a counting task},
  author={Gurnee, Wes and Ameisen, Emmanuel and Kauvar, Isaac and Tarng, Julius and Pearce, Adam and Olah, Chris and Batson, Joshua},
  journal={arXiv preprint arXiv:2601.04480},
  year={2026}
}

@misc{engels_not_2025,
    title = {Not {All} {Language} {Model} {Features} {Are} {One}-{Dimensionally} {Linear}},
    url = {http://arxiv.org/abs/2405.14860},
    doi = {10.48550/arXiv.2405.14860},
    urldate = {2025-04-16},
    publisher = {arXiv},
    author = {Engels, Joshua and Michaud, Eric J. and Liao, Isaac and Gurnee, Wes and Tegmark, Max},
    month = feb,
    year = {2025},
    note = {arXiv:2405.14860 [cs]},
}

@article{scikit-learn,
  title={Scikit-learn: Machine Learning in {P}ython},
  author={Pedregosa, F. and Varoquaux, G. and Gramfort, A. and Michel, V.
          and Thirion, B. and Grisel, O. and Blondel, M. and Prettenhofer, P.
          and Weiss, R. and Dubourg, V. and Vanderplas, J. and Passos, A. and
          Cournapeau, D. and Brucher, M. and Perrot, M. and Duchesnay, E.},
  journal={Journal of Machine Learning Research},
  volume={12},
  pages={2825--2830},
  year={2011}
}

@phdthesis{pedregosaizquierdo:tel-01100921,
  TITLE = {{Feature extraction and supervised learning on fMRI : from practice to theory}},
  AUTHOR = {Pedregosa-Izquierdo, Fabian},
  URL = {https://theses.hal.science/tel-01100921},
  NUMBER = {2015PA066015},
  SCHOOL = {{Universit{\'e} Pierre et Marie Curie - Paris VI}},
  YEAR = {2015},
  MONTH = Feb,
  TYPE = {Theses},
  HAL_ID = {tel-01100921},
  HAL_VERSION = {v2},
}

@inproceedings{rennie2005loss,
  title={Loss functions for preference levels: Regression with discrete ordered labels},
  author={Rennie, Jason DM and Srebro, Nathan},
  booktitle={Proceedings of the IJCAI multidisciplinary workshop on advances in preference handling},
  volume={1},
  pages={1--6},
  year={2005},
  organization={AAAI Press Menlo Park, CA}
}

@inproceedings{chen2016xgboost,
  title={Xgboost: A scalable tree boosting system},
  author={Chen, Tianqi and Guestrin, Carlos},
  booktitle={Proceedings of the 22nd acm sigkdd international conference on knowledge discovery and data mining},
  pages={785--794},
  year={2016}
}

@inproceedings{almasi-kristensen-mclachlan-2025-alignment,
    title = "Alignment Drift in {CEFR}-prompted {LLM}s for Interactive {S}panish Tutoring",
    author = "Almasi, Mina  and
      Kristensen-McLachlan, Ross Deans",
    editor = {Kochmar, Ekaterina  and
      Alhafni, Bashar  and
      Bexte, Marie  and
      Burstein, Jill  and
      Horbach, Andrea  and
      Laarmann-Quante, Ronja  and
      Tack, Ana{\"i}s  and
      Yaneva, Victoria  and
      Yuan, Zheng},
    booktitle = "Proceedings of the 20th Workshop on Innovative Use of NLP for Building Educational Applications (BEA 2025)",
    month = jul,
    year = "2025",
    address = "Vienna, Austria",
    publisher = "Association for Computational Linguistics",
    url = "https://aclanthology.org/2025.bea-1.6/",
    doi = "10.18653/v1/2025.bea-1.6",
    pages = "70--88",
    ISBN = "979-8-89176-270-1"
}

@inproceedings{
stolfo2025improving,
title={Improving Instruction-Following in Language Models through Activation Steering},
author={Alessandro Stolfo and Vidhisha Balachandran and Safoora Yousefi and Eric Horvitz and Besmira Nushi},
booktitle={The Thirteenth International Conference on Learning Representations},
year={2025},
url={https://openreview.net/forum?id=wozhdnRCtw}
}

@inproceedings{subramani-etal-2022-extracting,
    title = "Extracting Latent Steering Vectors from Pretrained Language Models",
    author = "Subramani, Nishant  and
      Suresh, Nivedita  and
      Peters, Matthew",
    editor = "Muresan, Smaranda  and
      Nakov, Preslav  and
      Villavicencio, Aline",
    booktitle = "Findings of the Association for Computational Linguistics: ACL 2022",
    month = may,
    year = "2022",
    address = "Dublin, Ireland",
    publisher = "Association for Computational Linguistics",
    url = "https://aclanthology.org/2022.findings-acl.48/",
    doi = "10.18653/v1/2022.findings-acl.48",
    pages = "566--581"
}

@inproceedings{hollinsworth-etal-2024-language,
    title = "Language Models Linearly Represent Sentiment",
    author = "Tigges, Curt  and
      Hollinsworth, Oskar J.  and
      Geiger, Atticus  and
      Nanda, Neel",
    editor = "Belinkov, Yonatan  and
      Kim, Najoung  and
      Jumelet, Jaap  and
      Mohebbi, Hosein  and
      Mueller, Aaron  and
      Chen, Hanjie",
    booktitle = "Proceedings of the 7th BlackboxNLP Workshop: Analyzing and Interpreting Neural Networks for NLP",
    month = nov,
    year = "2024",
    address = "Miami, Florida, US",
    publisher = "Association for Computational Linguistics",
    url = "https://aclanthology.org/2024.blackboxnlp-1.5/",
    doi = "10.18653/v1/2024.blackboxnlp-1.5",
    pages = "58--87"
}

@inproceedings{nguyen-etal-2025-multi,
    title = "Multi-Attribute Steering of Language Models via Targeted Intervention",
    author = "Nguyen, Duy  and
      Prasad, Archiki  and
      Stengel-Eskin, Elias  and
      Bansal, Mohit",
    editor = "Che, Wanxiang  and
      Nabende, Joyce  and
      Shutova, Ekaterina  and
      Pilehvar, Mohammad Taher",
    booktitle = "Proceedings of the 63rd Annual Meeting of the Association for Computational Linguistics (Volume 1: Long Papers)",
    month = jul,
    year = "2025",
    address = "Vienna, Austria",
    publisher = "Association for Computational Linguistics",
    url = "https://aclanthology.org/2025.acl-long.1007/",
    doi = "10.18653/v1/2025.acl-long.1007",
    pages = "20619--20634",
    ISBN = "979-8-89176-251-0"
}

@inproceedings{
li2023inferencetime,
title={Inference-Time Intervention: Eliciting Truthful Answers from a Language Model},
author={Kenneth Li and Oam Patel and Fernanda Vi{\'e}gas and Hanspeter Pfister and Martin Wattenberg},
booktitle={Thirty-seventh Conference on Neural Information Processing Systems},
year={2023},
url={https://openreview.net/forum?id=aLLuYpn83y}
}

@inproceedings{
marks2024the,
title={The Geometry of Truth: Emergent Linear Structure in Large Language Model Representations of True/False Datasets},
author={Samuel Marks and Max Tegmark},
booktitle={First Conference on Language Modeling},
year={2024},
url={https://openreview.net/forum?id=aajyHYjjsk}
}

@article{housen-kuiken-2009,
    author = { Alex Housen and Folkert Kuiken} ,
    title = {Complexity, Accuracy, and Fluency in Second Language Acquisition},
    journal = {Applied Linguistics},
    year =  {2009},
    vol = {30(4)},
    pages = {461–473}
}

@inproceedings{jin-etal-2026-toward,
    title = "Toward Beginner-Friendly {LLM}s for Language Learning: Controlling Difficulty in Conversation",
    author = "Jin, Meiqing  and
      Dugan, Liam  and
      Callison-Burch, Chris",
    editor = "Demberg, Vera  and
      Inui, Kentaro  and
      Marquez, Llu{\'i}s",
    booktitle = "Findings of the {A}ssociation for {C}omputational {L}inguistics: {EACL} 2026",
    month = mar,
    year = "2026",
    address = "Rabat, Morocco",
    publisher = "Association for Computational Linguistics",
    url = "https://aclanthology.org/2026.findings-eacl.47/",
    doi = "10.18653/v1/2026.findings-eacl.47",
    pages = "913--936",
    ISBN = "979-8-89176-386-9"
}

@article{Han_2024, 
    title={Chatgpt in and for second language acquisition: A call for systematic research}, 
    volume={46}, 
    DOI={10.1017/S0272263124000111}, 
    number={2}, 
    journal={Studies in Second Language Acquisition}, 
    author={Han, ZhaoHong}, 
    year={2024}, 
    pages={301–306}
}

@article{DEVORE2025101745,
title = {Exploring the ability of LLMs to classify written proficiency levels},
journal = {Computer Speech \& Language},
volume = {90},
pages = {101745},
year = {2025},
issn = {0885-2308},
doi = {https://doi.org/10.1016/j.csl.2024.101745},
url = {https://www.sciencedirect.com/science/article/pii/S0885230824001281},
author = {Susanne DeVore}
}

@inproceedings{hewitt-manning-2019-structural,
    title = "{A} Structural Probe for Finding Syntax in Word Representations",
    author = "Hewitt, John  and
      Manning, Christopher D.",
    editor = "Burstein, Jill  and
      Doran, Christy  and
      Solorio, Thamar",
    booktitle = "Proceedings of the 2019 Conference of the North {A}merican Chapter of the Association for Computational Linguistics: Human Language Technologies, Volume 1 (Long and Short Papers)",
    month = jun,
    year = "2019",
    address = "Minneapolis, Minnesota",
    publisher = "Association for Computational Linguistics",
    url = "https://aclanthology.org/N19-1419/",
    doi = "10.18653/v1/N19-1419",
    pages = "4129--4138"
}

@inproceedings{hewitt-liang-2019-designing,
    title = "Designing and Interpreting Probes with Control Tasks",
    author = "Hewitt, John  and
      Liang, Percy",
    editor = "Inui, Kentaro  and
      Jiang, Jing  and
      Ng, Vincent  and
      Wan, Xiaojun",
    booktitle = "Proceedings of the 2019 Conference on Empirical Methods in Natural Language Processing and the 9th International Joint Conference on Natural Language Processing (EMNLP-IJCNLP)",
    month = nov,
    year = "2019",
    address = "Hong Kong, China",
    publisher = "Association for Computational Linguistics",
    url = "https://aclanthology.org/D19-1275/",
    doi = "10.18653/v1/D19-1275",
    pages = "2733--2743"
}

@inproceedings{chi-etal-2020-finding,
    title = "Finding Universal Grammatical Relations in Multilingual {BERT}",
    author = "Chi, Ethan A.  and
      Hewitt, John  and
      Manning, Christopher D.",
    editor = "Jurafsky, Dan  and
      Chai, Joyce  and
      Schluter, Natalie  and
      Tetreault, Joel",
    booktitle = "Proceedings of the 58th Annual Meeting of the Association for Computational Linguistics",
    month = jul,
    year = "2020",
    address = "Online",
    publisher = "Association for Computational Linguistics",
    url = "https://aclanthology.org/2020.acl-main.493/",
    doi = "10.18653/v1/2020.acl-main.493",
    pages = "5564--5577"
}

@article{belinkov-2022-probing,
    title = "Probing Classifiers: Promises, Shortcomings, and Advances",
    author = "Belinkov, Yonatan",
    journal = "Computational Linguistics",
    volume = "48",
    number = "1",
    month = mar,
    year = "2022",
    address = "Cambridge, MA",
    publisher = "MIT Press",
    url = "https://aclanthology.org/2022.cl-1.7/",
    doi = "10.1162/coli_a_00422",
    pages = "207--219"
}

@inproceedings{
park2025the,
title={The Geometry of Categorical and Hierarchical Concepts in Large Language Models},
author={Kiho Park and Yo Joong Choe and Yibo Jiang and Victor Veitch},
booktitle={The Thirteenth International Conference on Learning Representations},
year={2025},
url={https://openreview.net/forum?id=bVTM2QKYuA}
}

@article{ramesh2022automated,
  title={An automated essay scoring systems: a systematic literature review},
  author={Ramesh, Dadi and Sanampudi, Suresh Kumar},
  journal={Artificial intelligence review},
  volume={55},
  number={3},
  pages={2495--2527},
  year={2022},
  publisher={Springer}
}

@article{shermis2014state,
  title={State-of-the-art automated essay scoring: Competition, results, and future directions from a United States demonstration},
  author={Shermis, Mark D},
  journal={Assessing Writing},
  volume={20},
  pages={53--76},
  year={2014},
  publisher={Elsevier}
}

@article{crossley2025large,
  title={A large-scale corpus for assessing source-based writing quality: ASAP 2.0},
  author={Crossley, Scott A and Baffour, Perpetual and Burleigh, L and King, Jules},
  journal={Assessing Writing},
  volume={65},
  pages={100954},
  year={2025},
  publisher={Elsevier}
}

@inproceedings{mathias2018asap++,
  title={ASAP++: Enriching the ASAP automated essay grading dataset with essay attribute scores},
  author={Mathias, Sandeep and Bhattacharyya, Pushpak},
  booktitle={Proceedings of the eleventh international conference on language resources and evaluation (LREC 2018)},
  year={2018}
}

@inproceedings{chi2026activations,
  title={Activations as Features: Probing LLMs for Generalizable Essay Scoring Representations},
  author={Chi, Jinwei and Wang, Ke and Chen, Yu and Lin, Xuanye and Xu, Qiang},
  booktitle={Proceedings of the AAAI Conference on Artificial Intelligence},
  volume={40},
  number={36},
  pages={30395--30403},
  year={2026}
}

@inproceedings{li2025kaes,
  title={KAES: Multi-aspect Shared Knowledge Finding and Aligning for Cross-prompt Automated Scoring of Essay Traits},
  author={Li, Xia and Pan, Wenjing},
  booktitle={Proceedings of the AAAI Conference on Artificial Intelligence},
  volume={39},
  number={23},
  pages={24476--24484},
  year={2025}
}

@inproceedings{do2023prompt,
  title={Prompt-and trait relation-aware cross-prompt essay trait scoring},
  author={Do, Heejin and Kim, Yunsu and Lee, Gary Geunbae},
  booktitle={Findings of the Association for Computational Linguistics: ACL 2023},
  pages={1538--1551},
  year={2023}
}

@inproceedings{chen2024plaes,
  title={Plaes: Prompt-generalized and level-aware learning framework for cross-prompt automated essay scoring},
  author={Chen, Yuan and Li, Xia},
  booktitle={Proceedings of the 2024 joint international conference on computational linguistics, language resources and evaluation (LREC-COLING 2024)},
  pages={12775--12786},
  year={2024}
}

@inproceedings{chen2023pmaes,
  title={PMAES: Prompt-mapping contrastive learning for cross-prompt automated essay scoring},
  author={Chen, Yuan and Li, Xia},
  booktitle={Proceedings of the 61st annual meeting of the association for computational linguistics (volume 1: long papers)},
  pages={1489--1503},
  year={2023}
}

@inproceedings{ridley2021automated,
  title={Automated cross-prompt scoring of essay traits},
  author={Ridley, Robert and He, Liang and Dai, Xin-yu and Huang, Shujian and Chen, Jiajun},
  booktitle={Proceedings of the AAAI conference on artificial intelligence},
  volume={35},
  number={15},
  pages={13745--13753},
  year={2021}
}

@article{ridley2020prompt,
  title={Prompt agnostic essay scorer: a domain generalization approach to cross-prompt automated essay scoring},
  author={Ridley, Robert and He, Liang and Dai, Xinyu and Huang, Shujian and Chen, Jiajun},
  journal={arXiv preprint arXiv:2008.01441},
  year={2020}
}

@article{santucci2020automatic,
  title={Automatic classification of text complexity},
  author={Santucci, Valentino and Santarelli, Filippo and Forti, Luciana and Spina, Stefania},
  journal={Applied Sciences},
  volume={10},
  number={20},
  pages={7285},
  year={2020},
  publisher={MDPI}
}

@inproceedings{forti2020malt,
  title={MALT-IT2: A new resource to measure text difficulty in light of CEFR levels for Italian L2 learning},
  author={Forti, Luciana and Bolli, Giuliana Grego and Santarelli, Filippo and Santucci, Valentino and Spina, Stefania},
  booktitle={Proceedings of the Twelfth Language Resources and Evaluation Conference},
  pages={7204--7211},
  year={2020}
}

@inproceedings{vajjala2014automatic,
  title={Automatic CEFR level prediction for Estonian learner text},
  author={Vajjala, Sowmya and Loo, Kaidi},
  booktitle={Proceedings of the third workshop on NLP for computer-assisted language learning},
  pages={113--127},
  year={2014}
}

@article{lagutina2024text,
  title={Text classification by CEFR levels using machine learning methods and the BERT language model},
  author={Lagutina, Nadezhda S and Lagutina, Ksenia V and Brederman, Anastasia M and Kasatkina, NN},
  journal={Automatic Control and Computer Sciences},
  volume={58},
  number={7},
  pages={869--878},
  year={2024},
  publisher={Springer}
}

@inproceedings{schmalz2021automatic,
  title={Automatic assessment of English CEFR levels using BERT embeddings},
  author={Schmalz, Veronica Juliana and Brutti, Alessio},
  booktitle={Proceedings of the Eighth Italian Conference on Computational Linguistics (CLiC-it 2021)},
  pages={295--301},
  year={2021}
}

@article{zhang2026automated,
  title={Automated text leveling for L2 English learners: a technology-enhanced framework with CEFR},
  author={Zhang, Xiaopeng and Zhang, Sumin},
  journal={Innovation in Language Learning and Teaching},
  pages={1--24},
  year={2026},
  publisher={Taylor \& Francis}
}

@inproceedings{ahlers2025classifying,
  title={Classifying German Language Proficiency Levels Using Large Language Models},
  author={Ahlers, Elias-Leander and Schilling, Malte},
  booktitle={2025 3rd International Conference on Foundation and Large Language Models (FLLM)},
  pages={638--645},
  year={2025},
  organization={IEEE}
}

@article{mizumoto2023exploring,
  title={Exploring the potential of using an AI language model for automated essay scoring},
  author={Mizumoto, Atsushi and Eguchi, Masaki},
  journal={Research Methods in Applied Linguistics},
  volume={2},
  number={2},
  pages={100050},
  year={2023},
  publisher={Elsevier}
}

@inproceedings{yancey2023rating,
  title={Rating short L2 essays on the CEFR scale with GPT-4},
  author={Yancey, Kevin P and Laflair, Geoffrey and Verardi, Anthony and Burstein, Jill},
  booktitle={Proceedings of the 18th workshop on innovative use of NLP for building educational applications (BEA 2023)},
  pages={576--584},
  year={2023}
}

@inproceedings{chang2022geometry,
  title={The geometry of multilingual language model representations},
  author={Chang, Tyler and Tu, Zhuowen and Bergen, Benjamin},
  booktitle={Proceedings of the 2022 Conference on Empirical Methods in Natural Language Processing},
  pages={119--136},
  year={2022}
}

@inproceedings{conneau2020emerging,
  title={Emerging cross-lingual structure in pretrained language models},
  author={Conneau, Alexis and Wu, Shijie and Li, Haoran and Zettlemoyer, Luke and Stoyanov, Veselin},
  booktitle={Proceedings of the 58th annual meeting of the association for computational linguistics},
  pages={6022--6034},
  year={2020}
}

@inproceedings{chi2020finding,
  title={Finding universal grammatical relations in multilingual BERT},
  author={Chi, Ethan A and Hewitt, John and Manning, Christopher D},
  booktitle={Proceedings of the 58th Annual Meeting of the Association for Computational Linguistics},
  pages={5564--5577},
  year={2020}
}

@article{skean2024does,
  title={Does representation matter? exploring intermediate layers in large language models},
  author={Skean, Oscar and Arefin, Md Rifat and LeCun, Yann and Shwartz-Ziv, Ravid},
  journal={arXiv preprint arXiv:2412.09563},
  year={2024}
}
\bibliographystyle{colm2026_conference}

\appendix
\section{Appendix}
\subsection{Language Model Specifications}
\begin{itemize}
\item The Qwen/Qwen3 Embedding Models of all sizes are described in \citet{qwen3embedding}. 
\item The google/embeddinggemma-300m descriptions are available at \citet{vera2025embeddinggemma}. 
\item The intfloat/multilingual-e5-large model is described in \citet{wang2024multilingual}, and the larger intfloat/e5-mistral-7b-instruct is described in \citet{wang2022text, wang2024improving}. 
\item The technichal report of nvidia/llama-embed-nemotron-8b can be found in \citet{babakhin2025llama}.
\end{itemize}

\subsection{Hyper Parameters of Probes}\label{appendix:parameters}
We have decided to include the hyper parameters used to fit the baseline model and probes in the appendix, as it was not a necessary inclusion for interpreting the results. The parameters is also available in our implementation of code.

\begin{itemize}
\item The XGBoost baseline model uses \texttt{reg:squared\_error} loss thereby optimizing for tree-based regression. It is initialized with 300 trees of depth 4 and a learning rate of 0.05. The remaining parameters is the defaults of \texttt{XGBRegressor} function from the python library xgboost \citep{chen2016xgboost}.

\item The Linear Regression model is a ridge regression with a $\alpha$ of 1, implemented in Scikit-Learn \citep{scikit-learn} with default parameters for the \texttt{Ridge} function. The model fits a parameter for all dimensions of the corresponding embedding model and a separate bias term. The resulting degrees of freedom is thus 1025 for Qwen3-0.6B, 2561 for Qwen3-4B, and 4097 for Qwen3-8B.

\item The Ordinal Regression model is a cumulative logistic link-function. The model is implemented with the \texttt{LogisticAT} function from the Mord library \citep{pedregosaizquierdo:tel-01100921}. 
The model fits a parameter for all dimensions of the corresponding embedding model and a separate bias term. This results in a latent space where the model fits four further thresholds for class separation. The resulting degrees of freedom is thus 1029 for Qwen3-0.6B, 2565 for Qwen3-4B, and 4102 for Qwen3-8B.

\item The logistic regression probe is fitted with the \texttt{LogisticRegression} function from Scikit-Learn. We changed \texttt{max\_iter=1000} from the default 100 to give the model more compute-time to converge. For the Qwen3-4B and Qwen3-8B model the embeddings the regression often failed to converged, but the predictive capabilities remained competitive with other linear probes, and was therefore kept in analysis.

\item The \texttt{MLPRegressor} function is initialized in Scikit-Learn with standard parameters except a extended \texttt{max\_iter=1000} to accommodate for the high-dimensional input data. The degrees of freedom scales quadratically with the size of corresponding models embedding-space. The resulting degrees of freedom is thus $(1024*1024+1)+(1024*1+1)=1.05M$ for Qwen3-0.6B, $6.56M$ for Qwen3-4B, and $16.78M$ for Qwen3-8B.

\item The \texttt{MLPClassifier} function is initialized in Scikit-Learn with similar parameters as the \texttt{MLPRegressor}, but has has five output neurons instead of one. The resulting degrees of freedom is thus $(1024*1024+1)+(1024*5+1)=1.05M$ for Qwen3-0.6B, $6.57M$ for Qwen3-4B, and $16.80M$ for Qwen3-8B.
The MLPClassifier is initialized with standard Scikit-Learn parameters with \texttt{max\_iter=1000}.

\end{itemize}

\subsection{Unclipped Linear Regression Residuals}\label{appendix:raw_residuals}
\begin{figure}[t]
\begin{center}
\includegraphics[width=0.95\linewidth]{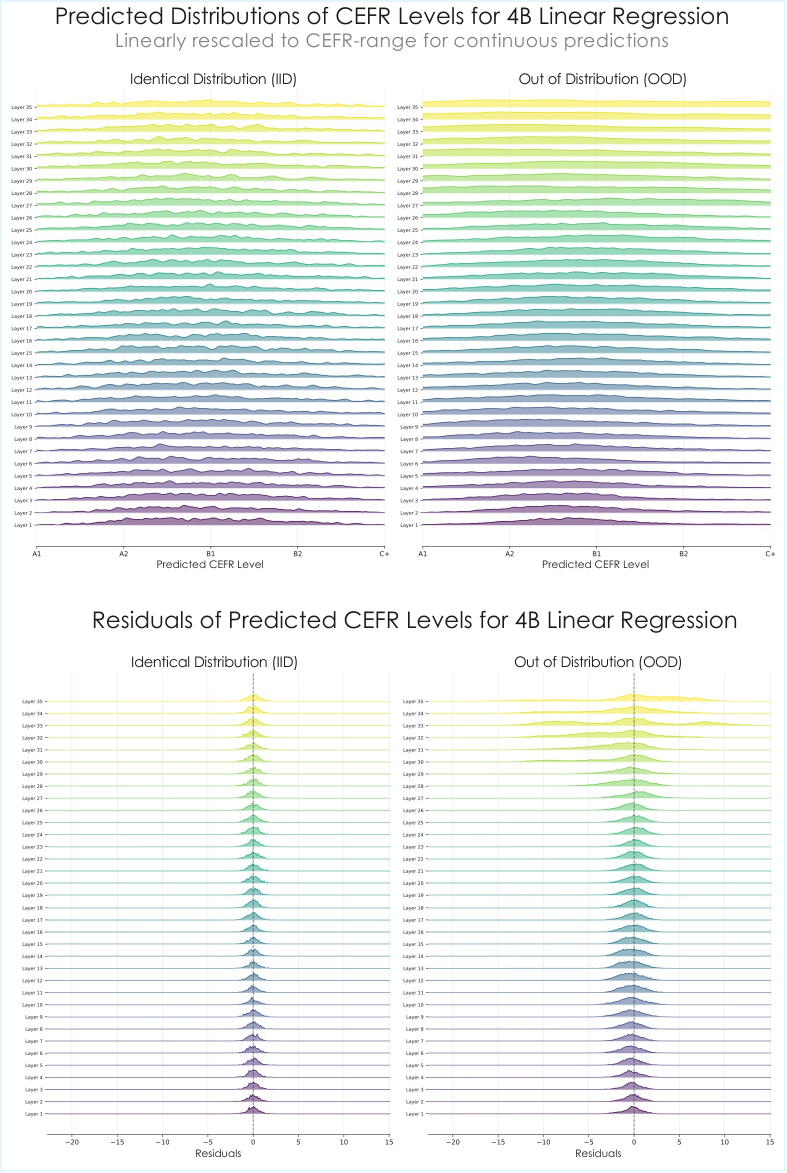}
\end{center}
\caption{Ridgeplot of development of prediction error. The upper two plots show the rescaled predictions
linear latent representation breaks by last layer.}
\label{fig:UnClip_Ridge}
\end{figure}

To further investigate the behavior of linear regression probe for Qwen3-4B, shown in \autoref{fig:Ridge}. We fit the linear regression probes for all 35 layers of Qwen3-4B and create four ridgeplots seen in \autoref{fig:UnClip_Ridge} showing the unclipped predictions and residuals. As \autoref{fig:Ridge} showed that the model quickly degrades to predicting outside of the labels-space, we utilize a linear rescaling instead of clipping to get predictions. The rescaling is unique for each probe, such that the highest value in a given layer becomes C+ and the lowest values becomes A1.

We see that the IID is able to retain a relatively tight residual distribution throughout all layers. This results in a jagged distribution resembling the original distribution. The OOD residuals however seem to rapidly flatten out around layer 27, resulting in very large residuals. The rescaled predictions of the model begins to degrade towards a uniform distribution of labels around layer 27 as a result.

\clearpage

\subsection{Probe Accuracy for Qwen3-8B}\label{appendix:Probes_Accuracy_8B}

We also include the probe accuracy for the 8B Qwen model-

\begin{figure}[htbp]
\begin{center}
\includegraphics[width=0.95\linewidth]{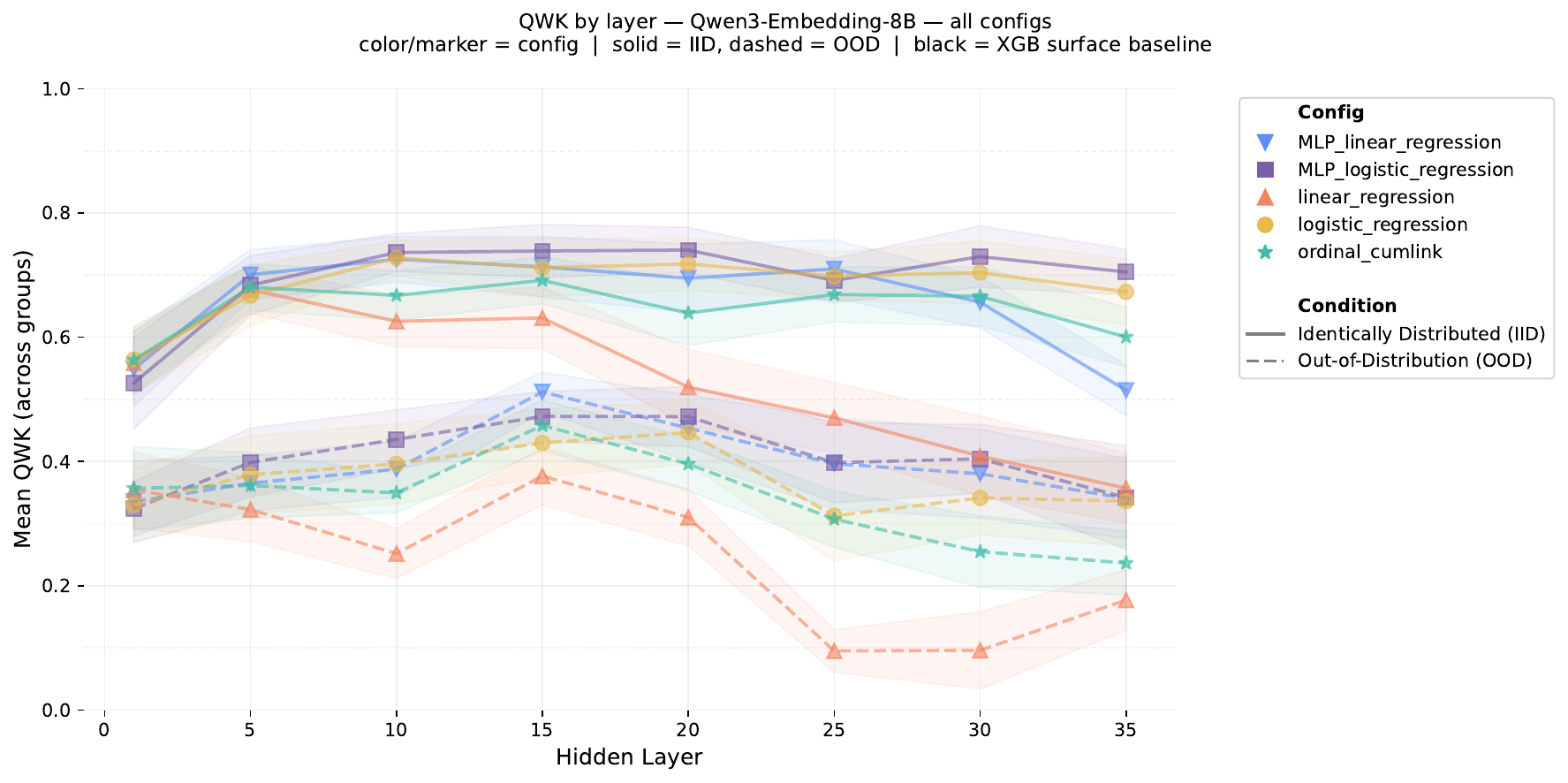}
\end{center}
\caption{Mean probe QWK across datasets for Qwen3-8B.}
\end{figure}
\end{document}